%% file: main.tex
\pdfoutput=1 %
\documentclass[letterpaper, 10 pt, conference]{ieeeconf}

\IEEEoverridecommandlockouts                               %

\usepackage{amsmath,amsfonts}
\allowdisplaybreaks
\usepackage{algorithmic}
\usepackage{algorithm}
\usepackage{array}
\usepackage{textcomp}
\usepackage{stfloats}

\usepackage{verbatim}
\usepackage{graphicx}

\usepackage{siunitx}
\usepackage{hyperref}
\usepackage{tikz}
\usepackage{placeins}

\usepackage[
 doi=false,isbn=false,url=false,eprint=false,
 backend=biber,
 style=ieee,
 bibstyle=ieee,
 citestyle=numeric-comp,
 sortcites=true,
 giveninits=true,
 maxcitenames=1,
 backref=false]{biblatex}
\AtEveryBibitem{%
  \clearfield{volume}%
  \clearfield{number}%
  \clearfield{pages}%
}
\DeclareRobustCommand{\citet}[1]{\citeauthor{#1}~\cite{#1}}
\DeclareRobustCommand{\citep}[1]{\cite{#1}}

\usepackage{trimclip}
\usepackage{adjustbox}
\usepackage{threeparttable} 
\usepackage{booktabs}
\usepackage{multirow}
\usepackage{array}
\usepackage{makecell}
\usepackage{caption}
\usepackage{subcaption}
\usepackage{tikz}
\usepackage{tabularx}
\let\labelindent\relax  %
\usepackage{enumitem}
\usepackage{tabularx}
\newcolumntype{Y}{>{\centering\arraybackslash}X}

\usepackage[dvipsnames]{xcolor}
\usepackage{dsfont} %
\usepackage{amssymb}
\usepackage{bm}
\usepackage{colortbl}
\usepackage{pdfrender}
\newcommand*{\boldcheckmark}{%
  \textpdfrender{
    TextRenderingMode=FillStroke,
    LineWidth=.5pt, %
  }{\checkmark}%
}

\definecolor{mynavy}{HTML}{1D6996}
\hypersetup{
    colorlinks=true,
    linkcolor=mynavy,
    citecolor=mynavy,
    filecolor=black,
    urlcolor=mynavy,
    pdftitle={Learning Where it Matters: Autonomous Data Collection and Offline Reinforcement Learning for Local Contact-Rich Manipulation},
    pdfauthor={Oliver Hausd\"orfer, Linus Schwarz, Gabor Marko, Christian Dietz, Timo Class, Luka Hofer, Jim Yun-Jin Li, Johannes Hechtl, Ralf R\"omer, Angela P. Schoellig},
}

\let\oldsubsection\subsection
\renewcommand{\subsection}{\vspace{-3pt}\oldsubsection}

\newcommand{\ieeetabstyle}{%
  \footnotesize
  \setlength{\tabcolsep}{0pt}%
  \renewcommand{\arraystretch}{1.15}%
}

\newcommand{\appendixrefname}{App.}
\newcommand{\appendixrefnamepl}{Apps.}
\DeclareRobustCommand{\appendixref}[1]{%
  \hyperref[#1]{\appendixrefname\nobreakspace\ref*{#1}}}
\DeclareRobustCommand{\appendixrefs}[2]{%
  \appendixrefnamepl\nobreakspace\hyperref[#1]{\ref*{#1}}~and~\hyperref[#2]{\ref*{#2}}}

\makeatletter
\@ifundefined{IEEEkeywords}{%
}{}
\makeatother

\begin{document}

\title{Learning Where it Matters: Autonomous Data Collection and Offline Reinforcement Learning for Local Contact-Rich Manipulation}

\author{Oliver Hausd\"orfer$^{1*}$, Linus Schwarz$^{1*}$, Gabor Marko$^{1}$, Christian Dietz$^{2}$, Timo Class$^{1}$,\\
Luka Hofer$^{1}$, Jim Yun-Jin Li$^{1}$, Johannes Hechtl$^{2}$, Ralf R\"omer$^{1}$, and Angela P. Schoellig$^{1}$%
\thanks{$^{*}$Equal contribution. $^{1}$The authors are with the Learning Systems and Robotics Lab, Technical University of Munich; and Munich Institute of Robotics and Machine Intelligence. This work was supported by the Robotics Institute Germany under BMBF grant 16ME0997K and by the Humboldt Professorship for Robotics and Artificial Intelligence. $^{2}$C.~Dietz and J.~Hechtl are with Siemens AG. Email: \href{mailto:oliver.hausdoerfer@tum.de}{oliver.hausdoerfer@tum.de}.}%
}

\maketitle

\begin{abstract}
Learned policies trained end-to-end on large datasets often remain brittle in high-precision contact-rich tasks. Further, they typically require large amounts of manually collected data and still struggle to generalize to scenes outside of the training scenarios. Classical motion planning, in turn, generalizes well but lacks the precision and contact awareness for tight-tolerance tasks. To reliably solve challenging contact-rich tasks, we propose to leverage robot data and learning where it matters in a modular approach. First, we introduce an autonomous data collection scheme for the real world that collects dense data locally in the contact-rich segment of the task, eliminating reliance on a teleoperator's skill. Second, we show that using offline RL, the collected data can be translated into policies with high success rates. We then integrate our local policies with traditional motion planning to solve long-horizon assembly tasks. Across four real-world tasks, using only \num{2}--\SI{2.5}{\hour} of autonomous data collection, we achieve an average success rate of \SI{96}{\percent} for long-horizon tasks, compared to \SI{55}{\percent} for a planning-only baseline and \SI{37}{\percent} for the strongest end-to-end baseline. Our local policies also reduce applied forces by \SI{49}{\percent}.
Notably, performance remains at \SI{90}{\percent} for scene-level out-of-distribution scenarios, while end-to-end approaches struggle. In total, we run over $1500$ real-world evaluation rollouts. Our results pave the way for targeted data collection in contact-rich tasks and for high success rates in precision applications. Videos and qualitative results are available at {\urlstyle{same}\url{https://learnsyslab.github.io/data_and_learning_where_it_matters/}}.

\end{abstract}

\section{Introduction}
\label{sec:introduction}

\begin{figure*}[!t]
    \centering
    \includegraphics[width=\textwidth]{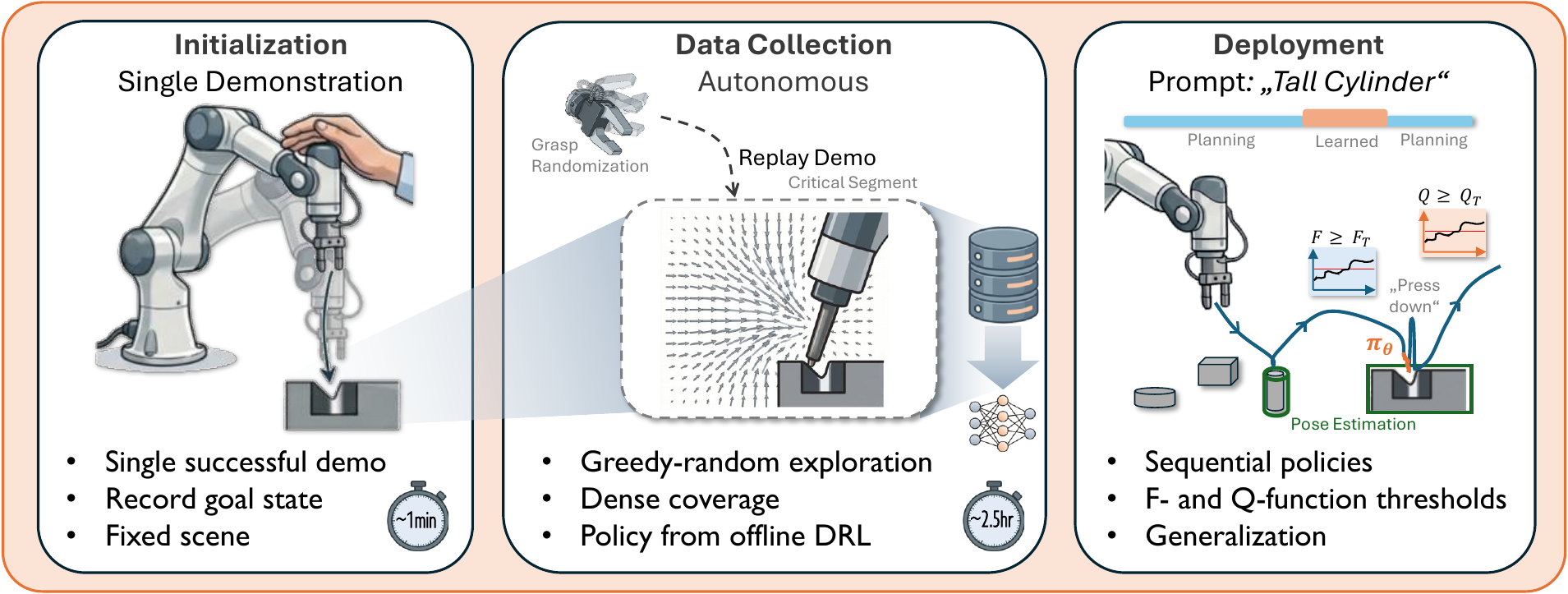}%
        \caption{ \textbf{Method}. We propose to learn local policies with offline deep reinforcement learning (DRL) using autonomously collected data, and integrate them with motion-planning to solve long-horizon tasks. (\textbf{Left}) First, we record a single demonstration in a fixed scene using kinesthetic teaching. The scene layout remains unchanged for the subsequent data collection, and the demonstration includes the scene reset. (\textbf{Middle}) The demonstration is replayed until the critical segment is reached, and then we execute a mixed greedy-random policy $\pi_{\text{explore}}$ for dense data collection at the insertion region, yielding successful and failed rollouts. We then use offline DRL to obtain a learned policy $\pi_\theta$ for the insertion. (\textbf{Right}) During deployment, $\pi_\theta$ for the contact-rich segment is sequentially chained with off-the-shelf pose estimation and motion planning to fulfill the task from object grasping to insertion.}
    \label{fig:method}
\end{figure*}

Contact-rich and high-precision tasks remain challenging for robot manipulation. Traditional methods relying on motion planning require accurate models of the contact dynamics, and pose estimation often lacks the precision required for tight tolerances~\cite{Morgan2021, fu20226droboticassemblybased, fu2023lanposelanguageinstructed6dobject}. At the same time, end-to-end robot learning has become a popular alternative for manipulation tasks~\cite{chi2024diffusionpolicyvisuomotorpolicy, intelligence2025pi05visionlanguageactionmodelopenworld, li2026forcevla2}, accelerated by large-scale data-collection efforts~\cite{o2024open, khazatsky2024droid}. Yet, end-to-end policies trained via imitation learning often struggle with low success rates and reliability in out-of-distribution scenarios. In particular, we observe that the policies typically fail in the same contact-rich, challenging task segments as traditional methods, indicating that end-to-end learning lacks structure or task composition to achieve a high success rate on these tasks.

Two key limitations of end-to-end learning approaches contribute to these failures. First, as we show, end-to-end policies rely heavily on the quality of teleoperated data. For challenging tasks, the operator needs training, and the quality of the resulting data varies with the operator's skill~\cite{belkhale2023dataqualityimitationlearning}. Second, end-to-end policies lack structure in data collection. Data is collected everywhere in the task instead of being targeted at the critical segment, leading to enormous data requirements. In addition, end-to-end policies are sensitive to changes in the scene, such as distractor objects or different backgrounds, which make it difficult to achieve high success rates reliably.

One way to make policies more structured is to follow a modular paradigm. Approaches that decompose tasks into modular segments have been proposed before~\cite{lee2020guideduncertaintyawarepolicyoptimization, sun2025arch, tang2024automatespecialistgeneralistassembly}. However, they have key limitations such as relying on online policy updates that lead to noisy learning~\cite{lee2020guideduncertaintyawarepolicyoptimization}, requiring a simulation setup~\cite{sun2025arch, tang2024automatespecialistgeneralistassembly}, or assuming perfect state knowledge during deployment~\cite{sun2025arch, tang2024automatespecialistgeneralistassembly} making them impractical for real-world applications.

To overcome the limitations, we propose an effective approach to collect data and learn policies for local contact-rich task segments in the real-world. In particular, we demonstrate that both autonomous data collection and offline RL are required to reliably achieve success rates of above \SI{90}{\percent} for local policies. Our scheme leverages \textit{data and learning where it matters}, eliminating both the constant dependency on a human teleoperator and noisy online policy updates. Our data collection targets the contact-rich task segment and requires only a single kinesthetic demonstration. We integrate learned local policies with motion planning to solve long-horizon tasks. For the other task segments, we rely on language-guided image segmentation~\cite{carion2026sam3segmentconcepts}, pose estimation~\cite{wen2024foundationposeunified6dpose}, and motion planning, none of which require robot-specific data.

Across four real-world tasks, our approach succeeds in \SI{96}{\percent} of trials after \num{2}--\SI{2.5}{\hour} of autonomous data collection, compared to \SI{55}{\percent} for a planning-only approach, and \SI{37}{\percent} for the strongest end-to-end baseline. Under scene-level distractors, its success rate remains between \SI{85}{\percent} and \SI{95}{\percent}, significantly improving upon end-to-end methods that are sensitive to scene-level changes, as we show.

\textbf{In summary, we present three core contributions.}

(\textbf{1}) We introduce an autonomous data-collection scheme for local contact-rich task segments, yielding $3$--$5\times$ more episodes than expert and novice teleoperators.

(\textbf{2}) We show that both our data-collection scheme and offline DRL improve the success rate on hardware compared to imitation learning and online DRL for local policies.

(\textbf{3}) We integrate our local policies with motion planning for long-horizon tasks using the Q-function as success predictor, achieving \SI{96}{\percent} success rate---outperforming end-to-end learning by +\SI{59}{\percent} and by +\SI{80}{\percent} when adding scene-level distractors.

\section{Related Work}
\label{sec:related_work}

\textbf{Data collection strategies.} Imitation learning typically relies on teleoperation data~\citep{intelligence2025pi05visionlanguageactionmodelopenworld, li2025trainrobotsimpactdemonstration, belkhale2023dataqualityimitationlearning}. DRL on hardware, for instance, via HIL-SERL similarly requires human intervention~\citep{luo2025precisedexterousroboticmanipulation}. Automated data-acquisition schemes are mostly limited to simulation. Assembly-by-disassembly generates demonstrations by automatic disassembly and trains a DRL policy using imitation rewards~\citep{tang2024automatespecialistgeneralistassembly}. Other approaches use search for data generation~\citep{ tian2024asapautomatedsequenceplanning} or to augment initial human demonstrations~\citep{ankile2024juicerdataefficientimitationlearning}. In contrast, we propose a practical automated data-acquisition scheme for the real world for contact-rich task segments.

\textbf{Contact-rich manipulation.} Methods relying on pure motion planning use fine-tuned pose estimation to achieve high success rates~\citep{fu20226droboticassemblybased, Morgan2021, fu2023lanposelanguageinstructed6dobject}. Learned policies achieve close to \SI{100}{\percent} when ground-truth states such as 6D object poses are available during deployment~\citep{tang2024automatespecialistgeneralistassembly,tang2023industrealtransferringcontactrichassembly,schoettler2020metareinforcementlearningroboticindustrial}, which limits real-world applicability. HIL-SERL demonstrated \SI{100}{\percent} success rate with DRL and vision input, but only for short-horizon tasks and depending on operator skill~\citep{luo2025precisedexterousroboticmanipulation,bi2023sampleefficientlearningsolverealworld}. Imitation policies typically achieve around \SI{80}{\percent} success rate on complex tasks~\citep{intelligence2025pi05visionlanguageactionmodelopenworld, ankile2024imitationrefinementresidual, goyal2024rvt2learningprecisemanipulation}, and we show that their failure cases are concentrated at critical segments. For traditional control methods, direct force or hybrid force-position control has been proven essential for contact-rich tasks~\citep{brown2026learninghybridcontrolpolicieshighprecision,shao2026interactiveforceimpedancecontrol}, with first learning-based models adopting the approach~\citep{li2026forcevla2,liang2022learningpreconditionshybridforcevelocity}.

\textbf{Modular Policies.} 
Many manipulation tasks can be separated into stages and sub-policies. GUAPO~\cite{lee2020guideduncertaintyawarepolicyoptimization} trains local vision-based policies in an online manner. However, GUAPO assumes pre-grasped parts, performs only insertion, and does not address when to stop the insertion or switch to a new task. Further, online policy updates are typically noisy. ARCH~\cite{sun2025arch} introduces a low-level primitive library chained by task-switching, but relies on state-conditioned policies and assumes perfect \textit{state knowledge} during deployment, significantly limiting real-world applicability. For robotic foundation models, mixture-of-experts architectures have become popular, enabling scalable multitask learning by specializing lightweight experts across behaviors~\citep {wang2024sparse, romer2026clare}. Switching happens through diffusion guidance~\citep{wang2024poco} or a learned router~\cite{chen2025multi}. In contrast to composing multiple learned policies, we exploit the structure of contact-rich manipulation itself, learning only the critical segment to improve success rate and scene-level OOD generalization. \autoref{tab:method_comparison} summarizes key differences between our work and related works.

\begin{table}[!t]
\centering
\caption{Related Work Comparison}
\label{tab:method_comparison}
\small
\setlength{\tabcolsep}{1.5pt} %
\setlength{\extrarowheight}{0pt} %
\begin{threeparttable}
\begin{tabularx}{\linewidth}{X c c c c c}
\toprule
\textbf{Properties} & GUAPO & ARCH & HIL-SERL & DP & Ours \\
\midrule
\makecell[l]{Requires state knowledge} & & \checkmark & & & \\
\arrayrulecolor{gray!30}\hline
Custom trained pose est.         & \checkmark\tnote{2} & & N/A & N/A & \\
\arrayrulecolor{gray!30}\hline
Pre-grasped parts       & \checkmark & & (\checkmark)\tnote{1} & & \\
\arrayrulecolor{gray!30}\hline
\makecell[l]{Requires pre-trained\\reward classifier} & (manual) & (sim) & \checkmark & N/A &  \\
\arrayrulecolor{black}\hline
Learns in real world           & \checkmark & (sim) &\checkmark &  \checkmark & \boldcheckmark \\
\arrayrulecolor{gray!30}\hline
\arrayrulecolor{gray!30}\hline
\arrayrulecolor{gray!30}\hline
Offline learning         & & (sim) & & \checkmark & \boldcheckmark\\
\arrayrulecolor{gray!30}\hline
Long-horizon tasks  & & \checkmark & & \checkmark & \boldcheckmark\\
\arrayrulecolor{gray!30}\hline
Data collection         & online & sim & onl./tele. & tele. & auto \\

\arrayrulecolor{black}\bottomrule
\end{tabularx}
\begin{tablenotes}[flushleft]
  \footnotesize
  \setlength{\itemsep}{0pt}
  \setlength{\parskip}{0pt}
  \setlength{\topsep}{0pt}
  \item[1] Partially: For all tasks, except USB-insertion and hand-over. 
  \item[2] Reported training effort: \SI{8}{\hour} on 4$\times$ P100 GPUs for a single object.
\end{tablenotes}
\end{threeparttable}
\end{table}

\section{Method}

\subsection{Problem Setup}
We model contact-rich interaction as an MDP $\mathcal{M} = (\mathcal{S}, \mathcal{A}, \rho, \mathcal{P}, r, \gamma)$, where $\mathbf{s}_t \in \mathcal{S}$ and $\mathbf{a}_t \in \mathcal{A}$ are the state and action at time step~$t$, $\rho(\mathbf{s}_0)$ is a distribution over initial states,  $\mathcal{P}(\mathbf{s}_{t+1} \mid \mathbf{s}_t, \mathbf{a}_t)$ are the unknown transition dynamics, $r_t=r(\mathbf{s}_t, \mathbf{a}_t) \in \mathbb{R}$ is a reward function encoding the task, and~$\gamma \in [0,1)$ is a discount factor. 
We observe the scene with a single wrist camera and obtain state estimates of task-relevant objects using pose estimation.
Our goal is to collect a dataset for training a policy $\pi_\theta: \mathcal{S} \rightarrow \mathcal{A}$ for the critical segment parameterized by a neural network with parameters~$\theta$ that maximizes the expected discounted return~$J(\pi_{\theta}) = \mathbb{E} \left[ \sum_{t=0}^{\infty} \gamma^t r_t \right]$. An overview of our methodology is provided in~\autoref{fig:method}. 

\subsection{Autonomous Offline Data Collection}
Our goal is to collect data for contact-rich tasks with minimal interventions to remove the dependency on a teleoperator. The idea is to set up a fixed scene with known states, and then conduct data collection within that scene. If necessary, fixtures can be 3D-printed to keep the objects in place. First, we record a single demonstration using kinesthetic teaching that performs the task and places all objects back to their initial positions. Our further scheme relies on keeping the scene layout unchanged during data collection.

We start by replaying the demonstration trajectory until the critical segment of the task is reached. In practice, the critical task segment is defined by the uncertainty of the pose estimation (see \hyperref[sec:app_pe]{Appendix}). Then, we execute a random-greedy data-collection policy

\begin{equation}
\label{eq:data_collection_policy}
\pi_{\text{explore}}(\mathbf{s}_t) = \mathbf{a}_t = \begin{cases} 
\mathbf{a} \sim \mathcal{U}(\mathcal{A})  & \text{with probability } p \\
\mathbf{a}^*_t & \text{with probability } 1-p,
\end{cases}
\end{equation}

where $\mathcal{U}$ is the uniform distribution, $p \in [0,1]$ is a probability threshold, and $\mathbf{a}_t^*$ is a greedy action. There are multiple ways to instantiate greedy policies~\citep{mandlekar2023mimicgendatagenerationscalable,mandlekar2023humaninthelooptaskmotionplanning,yeh2025actionconstrainedimitationlearning, tang2024automatespecialistgeneralistassembly}. We use the action that minimizes the distance to the goal state $\mathbf{s}_g$ with a velocity $v$:
\begin{equation}
\mathbf{a}^*_t = v \cdot \frac{ (\mathbf{s}_g + \Delta \mathbf{s}_{\text{grasp}}) - \mathbf{s}_t}{\|(\mathbf{s}_g + \Delta \mathbf{s}_{\text{grasp}}) - \mathbf{s}_t\|_2},
\end{equation} 
where $\Delta \mathbf{s}_{\text{grasp}}$ is the randomized grasp offset to the reference demonstration, and the state is the Cartesian end-effector pose. 
Additionally, we apply a safety filter to the action to prevent exploration from drifting outside the critical segment, constraining the end-effector position to a sphere with center $\mathbf{s}_\text{safe}$ and radius $r_\text{safe}$ during exploration, yielding the finally applied action: 

\begin{equation}
\mathbf{a}^{*\prime}_t =
\begin{cases}
v \cdot
\dfrac{\mathbf{s}_{\text{safe}} - \mathbf{s}_t}
{\|\mathbf{s}_{\text{safe}} - \mathbf{s}_t\|_2}
& \text{if }
\|\mathbf{s}_{\text{safe}} - \mathbf{s}_t\|_2> r_\text{safe}
\\
\mathbf{a}^*_t
& \text{otherwise}.
\end{cases}
\end{equation}

This promotes dense exploration within the contact-rich segment. Defining the safe exploration region requires the PE model’s
uncertainty, for which we collect 100 estimations within about
1 h using an automated script (\hyperref[sec:app_pe]{Appendix}). The data-collection episodes end either by truncation (timeout-based) or by task completion.

The task reward $r_t$ is a sparse reward determined by the distance of state $\mathbf{s}_t$ to the final state of the kinesthetic teaching demonstration $\mathbf{s}_g$ with reference to the grasp randomization $\Delta \mathbf{s}_{\text{grasp}}$: 

\begin{equation}
r_t = R_{\text{success}} \cdot \mathds{1} (\|\mathbf{s}_t - (\mathbf{s}_g + \Delta \mathbf{s}_{\text{grasp}})\|_2 \leq \epsilon).
\end{equation}

Computing the reward by state eliminates the need for a pre-trained vision-based reward classifier as required for other methods~\citep{lee2026roboreward, liang2026robometer}. As an advantage, after data collection, we obtain a vision-based classifier as the result of offline DRL in the form of the Q-function that we can use for downstream applications.

Contact-rich manipulation requires tight integration of perception and force feedback, particularly during the insertion phase~\citep{fang2026forcepolicylearninghybrid}. For this reason, we employ hybrid force-position control~\citep{brown2026learninghybridcontrolpolicieshighprecision, fang2026forcepolicylearninghybrid} during the insertion phase.

\subsection{Offline DRL for Policy Learning}

To learn from few demonstration trajectories, we require a sample-efficient learning algorithm. Imitation learning is not sufficient, since we need to learn from suboptimal and failed trajectories. Thus, we use offline DRL. Similar to RLPD~\citep{ball2023efficientonlinereinforcementlearning}, we build on top of SAC~\citep{haarnoja2018softactorcriticoffpolicymaximum} and employ improvements for sample efficiency. To mitigate Q-function overestimation, we add layer normalization~\citep{ba2016layernormalization} to the critic and use Randomized Ensembled Double Q-Learning~\citep{chen2021randomizedensembleddoubleqlearning}, where we train $N_Q$ Q-functions and select a random subset $S$ of size $N_{Q,S}$ to compute the TD-targets as

\begin{equation}
    y_t = r_t + \gamma \mathbb{E}_{\mathbf{a}' \sim \pi_\theta} \left[ \min_{i \in S} Q_{\bar{\phi}_i}(\mathbf{s}_{t+1}, \mathbf{a}') - \alpha \log \pi_\theta(\mathbf{a}' \mid \mathbf{s}_{t+1}) \right].
\end{equation}

The Q-functions and policy are trained via the standard SAC objective~\citep{haarnoja2018softactorcriticoffpolicymaximum} using $y_t$. Since our setting does not rely on online learning, we fix the entropy coefficient $\alpha$ rather than adapting it during learning as typically done for online methods.%

\subsection{Policy Integration in Long-Horizon Tasks}
Previous approaches propose local policies, but do not consider their integration in long-horizon tasks~\cite{lee2020guideduncertaintyawarepolicyoptimization}. In our case, interaction with the scene is done using a language-guided segmentation model (SAM3~\citep{carion2026sam3segmentconcepts}), which we prompt with the commands ``purple Lego brick'', ``gray plastic cover with circular grille'', and ``yellow cardboard salt box'' for our tasks. We feed the segmentation masks into an off-the-shelf pose estimation model~\citep{wen2024foundationposeunified6dpose, ippe_pe} to obtain 6D pose estimates for the two objects in the scene that we manipulate, unless otherwise specified. Pose estimation is run multiple times to refine the estimates throughout the trajectory (\hyperref[sec:app_pe]{Appendix}). The pose estimates serve as the basis for motion planning, which uses simple waypoint planning and following. Our formulation can easily be extended to include trajectory optimization and obstacle avoidance.

All waypoints for motion planning are defined relative to the object's 6D pose and based on the single demonstration. The robot first moves towards the pick-up object until it reaches a specified waypoint. Then, we grasp the object and transport it to the critical segment. The learned insertion policy is triggered once contact is made, i.e., if \mbox{$\|\mathbf{F}\|_2 \geq \mathcal{F}_{\text{threshold}}$}, which we found the most reliable trigger similar to other works~\citep {stranghoener2026sharerlstructuredinteractivereinforcement,zhang2021learninginsertionprimitivesdiscretecontinuous}.

Then, the learned policy is executed and terminated based on the learned Q-functions and a threshold $\lambda_{\text{success}}$ to detect successful insertion:

\begin{equation}
    Q(\mathbf{s}_t, \pi_\theta(\mathbf{s}_t)) = \frac{1}{N_Q} \sum_{i=1}^{N_Q} Q_{\phi_i}(\mathbf{s}_t, \pi_\theta(\mathbf{s}_t)) > \lambda_{\text{success}},
\end{equation}
We show the resulting Q-value predictions $Q_{\phi}(\mathbf{s}_t, \pi_{\theta}(\mathbf{s}_t))$ as a function of policy timesteps for truncated and successful trajectories in \autoref{fig:q_function_success_classifier}, where the separation between successful and truncated trajectories is clearly visible.

After the learned policy terminates, any remaining motions, such as a push-down to complete the insertion, are executed.

\input{task_images}

\section{Experimental Results}
\label{sec:result}

\subsection{Task descriptions}
We evaluate our approach on real-world contact-rich tasks that require handling deformable components and precise alignment (\autoref{fig:tasks_images}). We define \textit{partial success} for all tasks, indicating correct coarse placement but failed insertion.

\textbf{Shelf stocking}: This task requires stocking a tightly packed shelf with a yellow salt box. Most objects on the shelf are made of cardboard and may deform during interaction. The policies need to learn to react to those changes, precisely align the salt box, and perform an additional push to fully complete the insertion. \textit{Partial success} is if the object is placed on the shelf, but not inserted.

\textbf{LEGO stacking}: The task is to pick up a brick and stack it on another $4\times2$ brick from an unknown position on the base plate. The task is solved once the two bricks are fully stacked together. \textit{Partial success} is defined as the blocks being placed on top of each other, but not fully inserted.

\textbf{Fan cover}: This task originates from a real-world production line that currently necessitates manual assembly. The assembly requires aligning the studs through intricate, contact-rich motions against the base. The injection-molded PBT/PC parts deform during assembly. \textit{Partial success}: the cover is placed on the base, but not fully inserted.

\textbf{Fan cover (hard)}: A hard version of this task introduces additional diversity during data collection and deployment.

\subsection{Implementation Details}
\textbf{Baselines.} For imitation learning we use DiTFlow~\citep{dasari2024ingredientsroboticdiffusiontransformers} and Diffusion Policy (DP)~\citep{chi2024diffusionpolicyvisuomotorpolicy} as task-specific policies that we train from scratch, and $\pi_{0.5}$~\citep{intelligence2025pi05visionlanguageactionmodelopenworld} for a foundation model that we fine-tune. For DiTFlow, we include a version trained on data collected by a novice teleoperator. We use HIL-SERL as baseline for online DRL with human intervention~\citep{luo2025precisedexterousroboticmanipulation}. We also test against pure pose estimation and motion planning (PE \& MP). For the local policies, we compare against GUAPO, which uses online learning and distance-based switching.

\begin{table}[!t]
\centering
\caption{Data collection and training hyperparameters}
\label{tab:sac_hyperparams}
\ieeetabstyle %
\renewcommand{\arraystretch}{1.0}%
\setlength{\tabcolsep}{3pt}%
\newcommand{\grouplabel}[1]{\multicolumn{3}{@{}l}{\itshape #1} \\}%
\begin{tabularx}{\columnwidth}{@{}l l >{\raggedright\arraybackslash}X@{}}
\toprule
\textbf{Parameter} & \textbf{Value} & \textbf{Explanation} \\
\midrule
\grouplabel{Data collection}
$p$ & 0.8 & Probability of choosing a random action. \\
$L_\text{trunc}$ & 150 & Episode truncation length (steps). \\
$R_\text{success}$ & 10 & Task completion reward. \\
\midrule
\grouplabel{Policy}
Batch size & 512 & Samples per optimization step. \\
Q lr & $5 \times 10^{-4}$ & Adam step size for critic ensemble. \\
Policy lr & $3 \times 10^{-4}$ & Adam step size for actor. \\
$\alpha$ & $1 \times 10^{-3}$ & Fixed entropy regularization weight. \\
$\gamma$ & 0.97 & Reward discount factor. \\
$\tau$ & 0.999 & Polyak averaging coefficient. \\
UTD & 5 & Update-to-data ratio, equals num. epochs. \\
$N_Q$ / $N_{Q,S}$ & 10 / 2 & Critic ensemble size / subset size. \\
Actor & MLP & Two layers, hidden size 128, ReLU. \\
Q-functions & MLP & Two layers, hidden size 128, LayerNorm, ReLU. \\
\midrule
\grouplabel{Control}
$f_\text{low-level}$ & \SI{1000}{\hertz} & Cartesian/torque controllers~\citep{pro2025crispcompliantros2}. \\
$f_\text{policy}$ & \SI{15}{\hertz} & Control frequency of the DRL policy. \\
\bottomrule
\end{tabularx}
\end{table}

\textbf{Hardware.} We use a Franka Research 3 with a BOTA-DENS-IND2-B4 FT-Sensor and a wrist-mounted Intel RealSense D405 camera. End-to-end policies additionally use an external camera. 

\begin{figure}[!b]
    \centering
    \includegraphics[width=\columnwidth]{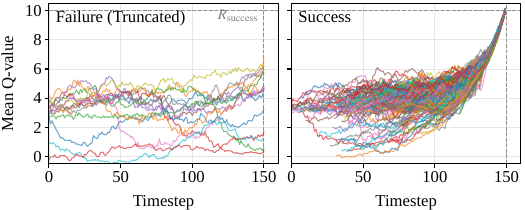}%
    \caption{\textbf{Q-function success classifier for real-world rollouts.} Q-function predictions for 300 real-world rollouts of which 285 are successful (\SI{95}{\percent} SR). Successful episodes are aligned such that successful termination occurs at timestep $t = 150$. The maximum Q-value is determined by the sparse reward $R_\text{success} = 10$.}
    \label{fig:q_function_success_classifier}
\end{figure}

\textbf{Training}. Our policy is trained for roughly \SI{1}{\hour} on a local workstation with an NVIDIA RTX 5090 GPU. We train DP and DiTFlow on the same workstation for \SI{5}{\hour} and 50k steps, and $\pi_{0.5}$ is fine-tuned for 40k steps using 8$\times$H100 GPUs for \SI{18}{\hour}. Training parameters are given in \autoref{tab:sac_hyperparams}. For data collection, we set the probability of random actions to $p = 0.8$ and truncate episodes after \SI{10}{\second} (150 steps). During deployment, we use $\lambda_{\text{success}}=0.93 R_\text{success}$ for the Q-function-based success classifiers (\autoref{fig:q_function_success_classifier}). The policy receives a single RGB viewpoint from a wrist-mounted camera, and pose estimation uses additional depth measurements~\citep{wen2024foundationposeunified6dpose, ippe_pe}. The images are embedded using DINOv2~\citep{caron2021emerging}, and projected down to 16 dimensions using an MLP before concatenation with the state. The state includes F/T measurements, controller error, Cartesian velocity, and the previous action.

\input{table_time_vertical}
\vspace{1mm}

\subsection{Can We Efficiently Collect Data in The Real World at The Critical Segment?}
We compare our scheme against teleoperation of a novice and expert operator, aiming for equal wall-clock time in all cases (\autoref{tab:data_efficiency}). During the same wall-clock time, our data-collection scheme collects significantly more data at the critical segment, resulting in $3$--$5\times$ more collected episodes. The majority of the time saving compared to teleoperations comes from the fact that the scene does not need to be manually reset between episodes. Further, for difficult tasks the number of failed teleoperation episodes increases, putting significant strain on the operator. Our data-collection scheme only requires interventions when the scene is not reset correctly, as shown in suppl. video and on website.

\begin{table}[!b]
\centering
\caption{Data $\times$ Policy Combinations}
\label{tab:data_policy}
\small
{\setlength{\tabcolsep}{8pt}%
\begin{threeparttable}
\begin{tabularx}{\linewidth}{>{\centering\arraybackslash}X c c c}
\multicolumn{4}{c}{Hardware experiments (LEGO task). SR (partial SR [\%]).}\\
\toprule
\makecell[c]{\textbf{Teleop Data}\\\textbf{+ Offline DRL}} & \makecell[c]{\textbf{Our Data}\tnote{1}\\\textbf{+ DP}} & \makecell[c]{\textbf{Our Data}\\\textbf{($\bm{p}\!=\!\bm{0}$)\tnote{2}+ DP}} & \textbf{Ours} \\
\midrule
30 (64) & 20 (74) & 84 (100) & \textbf{94 (100)} \\
\bottomrule
\end{tabularx}
\begin{tablenotes}[para,flushleft]
  \footnotesize
  \setlength{\itemsep}{0pt}
  \setlength{\parskip}{0pt}
  \setlength{\topsep}{0pt}
  \item[1] Same data as \textbf{ours}, restricted to successful episodes (375/500).
  \item[2] To obtain data suited for behavior cloning (BC), we record for another \qty{2.5}{\hour} using perfect actions without exploration ($p{=}0$). With our offline DRL, this data yields \SI{20}{\percent} (\SI{80}{\percent}) (not shown in table), matching our sim results (\autoref{fig:results_datacollection}, middle).%
\end{tablenotes}
\end{threeparttable}}

\end{table}

\begin{figure}[!t]
    \centering
    \includegraphics[width=\columnwidth]{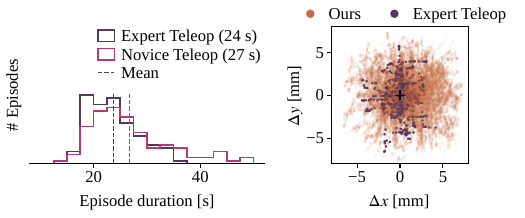}%
    \caption{\textbf{Data Collection.} (\textbf{Left}) Distribution of episode durations for expert and novice teleoperators. High-quality data is characterized by uniform and short trajectories. (\textbf{Right}) Distribution of states in the critical segment. Our data-collection schemes yield dense and uniform coverage.}
    \label{fig:dc_overview}
\end{figure}

\autoref{fig:dc_overview} shows the data distribution in the critical segment. Our approach uniformly covers the critical segment, while teleoperation covers it only sparsely. We further find that demonstration quality depends on operator skill. Demonstrations from a novice teleoperator are characterized by overall longer trajectories (\SI{27}{\second} vs. \SI{24}{\second} on average). They are often multi-modal, i.e., the task is demonstrated in different ways, and exhibit higher speed variance  ($\mathrm{CV}_{\text{novice}} = 0.285$ vs. $\mathrm{CV}_{\text{expert}} = 0.185$). This data quality difference is reflected in policy success rates (\autoref{tab:sr_results}). 

\begin{figure}[!t]
    \centering
    \includegraphics[width=\columnwidth]{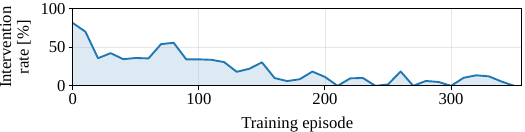}%
    \caption{\textbf{HIL-SERL training.} Human intervention rate over training episodes for the LEGO insertion phase on hardware, averaged over buckets of 10 episodes. HIL-SERL requires 238 teleoperation interventions throughout training. At the end of training HIL-SERL reached \SI{92}{\percent} SR.}
    \label{fig:hilserl_interventions}
\end{figure}

\begin{table}[!b]
\centering
\caption{Method Comparison}
\label{tab:method_ablation}
\small
{\setlength{\tabcolsep}{3pt}%
\begin{tabularx}{\linewidth}{*{3}{>{\centering\arraybackslash}X}}
\multicolumn{3}{c}{Sim. ablations (LEGO task). SR [\%] ($\pm$ std, 3 seeds).}\\
\toprule
\textbf{\makecell{Dist. Switch\\(cf. GUAPO)}} & \textbf{\makecell{Online DRL\\(cf. GUAPO)}} & \textbf{\makecell{Ours\\\vphantom{Control}}} \\
\midrule
94.9 & 95.5 $\pm$ 5.0  & \textbf{98.3 $\pm$ 0.6} \\
\bottomrule
\end{tabularx}}
\end{table}

\vspace{1mm}

\subsection{Is Offline DRL Required to Learn Policies With High Success Rates?}
We compare offline DRL, online DRL, and imitation learning using both teleoperated data and our data-collection scheme. All are local policies under the same PE \& MP wrapper. The combination of offline DRL with our collected data is key for a high success rate (\autoref{tab:data_policy}). Our data-acquisition scheme yields dense data coverage with suboptimal trajectories, and offline DRL can leverage them for policy learning. Our data-acquisition scheme using $p=0$ (only optimal trajectories) also achieves a high SR of \SI{84}{\percent} when combined with imitation learning (DP). Data from the human teleoperator, however, only yields \SI{30}{\percent} due to limited data coverage and diversity, showing that reliable and automated data collection is key for high success rates.

Using online DRL yields lower SR, and the results are dependent on the training seed. This may necessitate multiple training runs on hardware to learn a reliable policy. We also run HIL-SERL for the critical segment (\autoref{fig:hilserl_interventions}) and achieve \SI{92}{\percent} (\SI{92}{\percent}) SR, requiring a total of 238 human teleoperation takeovers in 160 episodes (356 training episodes in total). We warm-start HIL-SERL from 20 teleoperation demos as in the original implementation. The dependency on human interventions and high variance in online training are also observed in other works~\citep{chen2025conrftreinforcedfinetuningmethod, lu2025hiors}. Changing to GUAPO's distance-based policy switching yields slightly lower SR compared to our contact switch (\SI{-3.4}{\percent}).

\input{table_main_results}

\vspace{1mm}
\subsection{Can Our Local Policies be Integrated in Long-Horizon Tasks, and Does Success Rate Remain High?}\label{sec:results_main}

We integrate our policies with PE \& MP for long-horizon tasks---including picking up objects and inserting them---and compare against end-to-end learning.

Our method combining learned policies with PE \& MP achieves \SI{96}{\percent}, compared to the strongest baseline at \SI{55}{\percent}, and improves on average by +\SI{59}{\percent} over baselines. Grasping and free-space motion are already successfully solved by PE \& MP (\SI{97}{\percent} SR), while our local policies add the skill to perform the insertions reliably. In effect, the local policy corrects residual pose estimation errors that prevents PE \& MP from completing the task on its own, leveraging contact information in the process. We show remaining failure cases of our method  in~\autoref{fig:failure_cases}.

\begin{figure}[!b]
    \centering
    \includegraphics[width=\columnwidth]{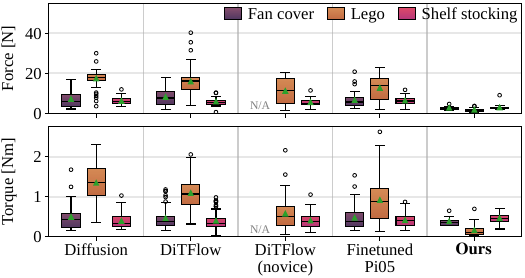}%
    \caption{\textbf{Contact forces and torques.} Maximum forces (\textbf{top}) and torques (\textbf{bottom}) applied during policy rollouts. \textbf{Ours} applies substantially lower forces and torques to the objects than the baselines.}
    \label{fig:force_results}
\end{figure}

Our policies also reduce the forces applied to the objects, which is critical for sensitive parts (\autoref{fig:force_results}). Applied maximum forces are reduced on average by \SI{49}{\percent} (\SI{-4.7}{\newton}). This is possible, because we have a separate push-down motion with the force limited to the maximum force applied during data collection. Kinesthetic teaching naturally gives the operator more precise control of the robot compared to teleoperation, thus allowing for more dexterous insertion.%

End-to-end methods capture the task semantics and get coarse placement right, but struggle at the actual challenging insertion phase. This is reflected in high \textit{partial} success rates (\SI{95}{\percent} for DP) that only measure coarse placement. The policies then often fail at the challenging section, as shown by low overall success rates (\SI{36}{\percent} for DP) (\autoref{tab:sr_results}). HIL-SERL struggles with long task horizons, leading to low success rates. Primarily, the random gripper actions during exploration lead to frequent task failures when already grasped objects are dropped, and the task needs to be reset.

PE \& MP outperforms end-to-end learning. Pose estimation alone is precise enough to solve the tasks \SI{55}{\percent} of the time. While pose estimation is good enough to achieve grasping and coarse placement, it lacks the precision to reliably achieve tight insertions. Notably, PE \& MP does not require any robot-specific data at all and still outperforms end-to-end learning in our settings.

In the suppl. video and website, we show further long-horizon tasks.

\vspace{1mm}
\subsection{How Do End-to-End Policies and Our Method Generalize to Scene-Level Distractors?}

\begin{figure}[!b]
    \centering
    \noindent
    \begin{tikzpicture}
      \path[use as bounding box] (0,0) rectangle (0.49\columnwidth, 3cm);
      \clip (0,0) rectangle (0.49\columnwidth, 3cm);
      \node[anchor=south west, inner sep=0] at (0, 0) {%
        \includegraphics[height=4cm]{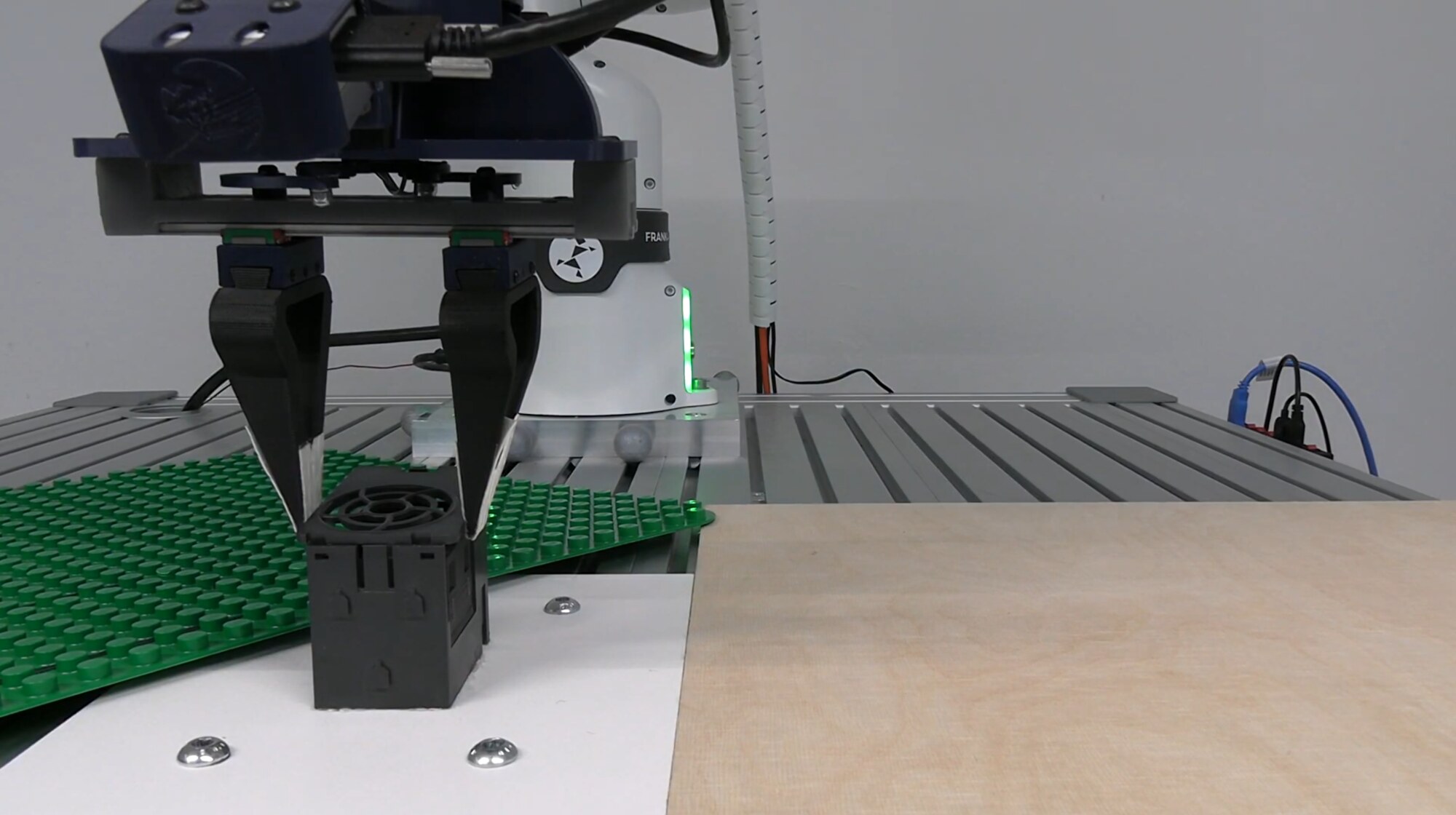}%
      };
    \end{tikzpicture}%
    \hfill
    \begin{tikzpicture}
      \path[use as bounding box] (0,0) rectangle (0.49\columnwidth, 3cm);
      \clip (0,0) rectangle (0.49\columnwidth, 3cm);
      \node[anchor=north west, inner sep=0] at (-0.1103\columnwidth, 3cm) {%
        \includegraphics[width=0.6738\columnwidth]{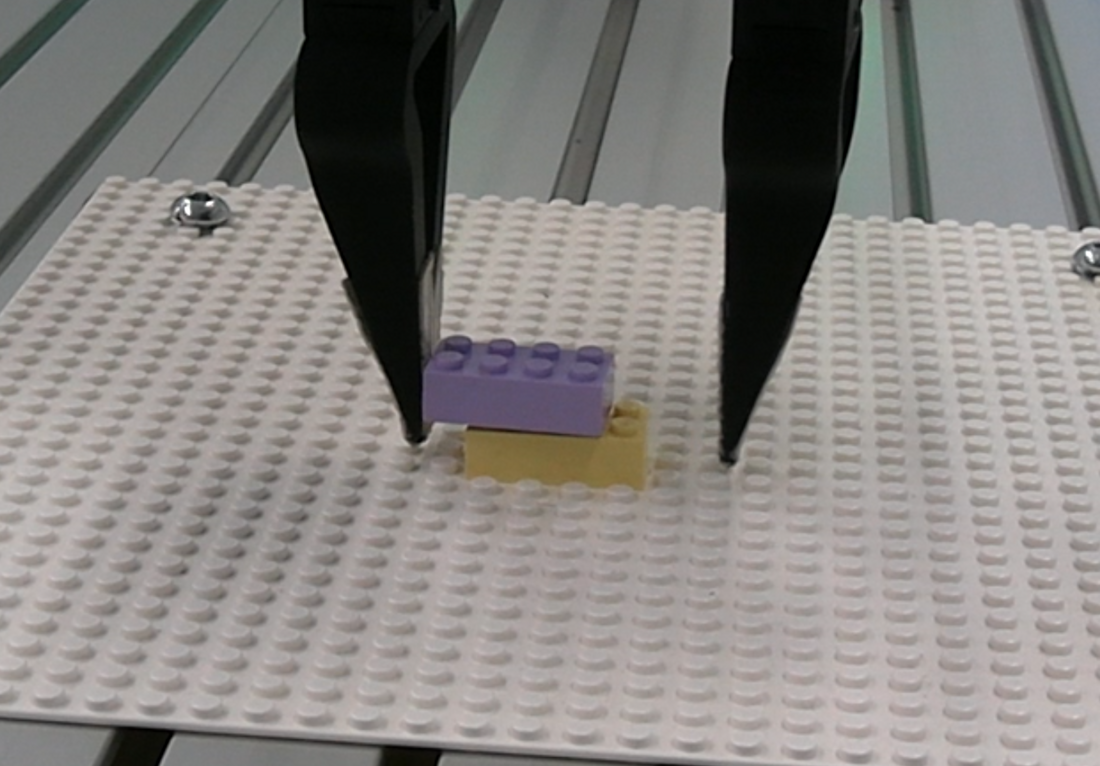}%
      };
    \end{tikzpicture}
    \caption{\textbf{Failure cases.} (\textbf{Left}) Our policy fails if distractors are placed very close to the object within the cropped region of the wrist image. (\textbf{Right}) Hardware limitations, e.g., objects sticking to the 3D-printed gripper.}
    \label{fig:failure_cases}
\end{figure}

During deployment, policies should operate under various conditions, including scene-level changes such as background changes and distractor objects. We expose our method to various OOD scenarios, including different object configurations, distractor objects, and novel backgrounds (\autoref{fig:tasks_images}). For our method, the success rates stay close to the in-distribution results (\SI{-6}{\percent}), as the learned policies operate locally and are largely unaffected by the changes. Occurring failure cases include distractor objects being placed close to the insertion point, taking local visual observations out of distribution (\autoref{fig:failure_cases}). We present additional results for object avoidance, dynamically grasping from a human hand, arbitrary object poses, and further randomizations---all without additional data collection---in the supplementary video.

\begin{figure*}[!t]
    \centering
    \noindent
    \begin{tikzpicture}
      \path[use as bounding box] (0,0) rectangle (\linewidth, 3.555cm);
    
      \node[anchor=center, inner sep=0] at (0.19\linewidth, 1.755cm) {%
        \includegraphics[height=0.1755\linewidth]{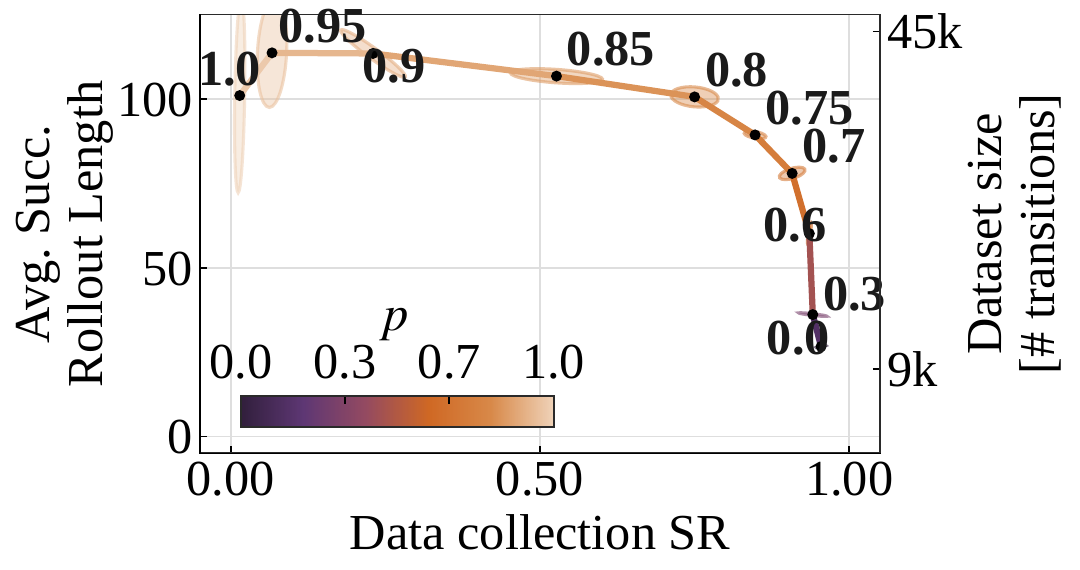}%
      };
      \node[anchor=center, inner sep=0] at (0.545\linewidth, 1.755cm) {%
        \includegraphics[height=0.1755\linewidth]{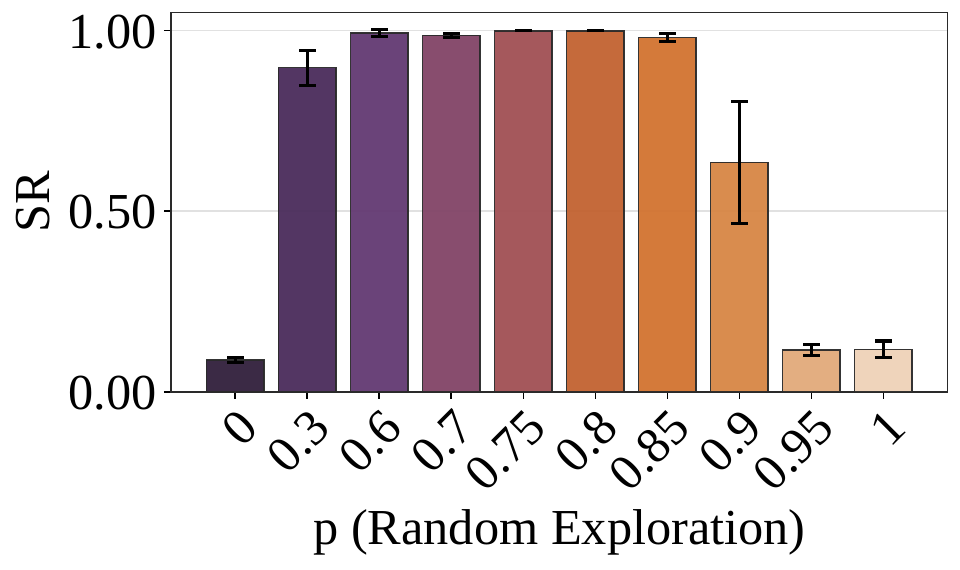}%
      };
      \node[anchor=center, inner sep=0] at (0.855\linewidth, 1.755cm) {%
        \includegraphics[height=0.1755\linewidth]{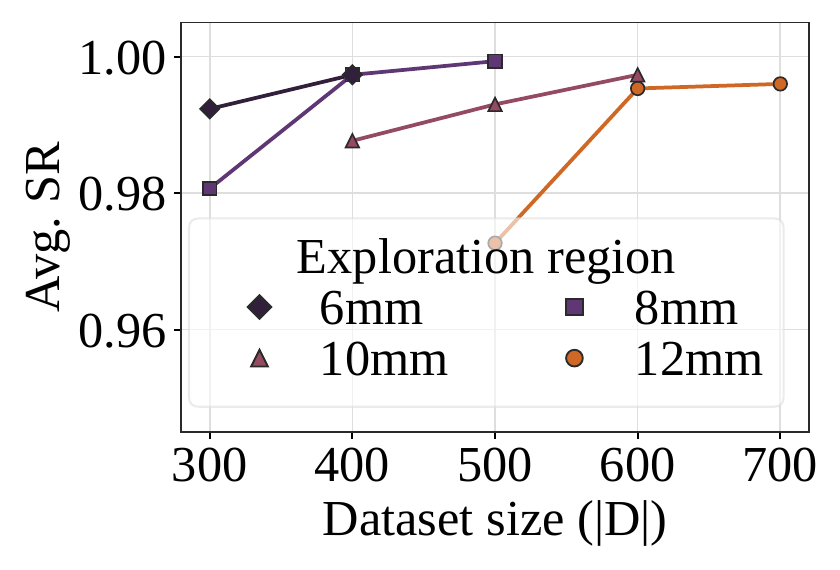}%
      };
    \end{tikzpicture}
    \vspace{-23pt}
    \caption{\textbf{Data-collection ablations.} Ablation of data-collection parameters. \textbf{(Left)} Effect of random action sampling rate $p$ on average successful rollout length, dataset size, and success rate during data collection. \textbf{(Middle)} Trained policy success rate for different random exploration rates $p$ during data collection. \textbf{(Right)} Scaling of SR and dataset size with pose estimation uncertainty.}
    \label{fig:results_datacollection}
\end{figure*}

For end-to-end methods, SRs decline by \SI{-24}{\percent} (\mbox{\autoref{tab:sr_results_generalization}}), showing---at least for our data regime---that the policies do not generalize well. The fan cover (hard) task includes the standard fan cover OOD scenarios during training, and during evaluation, additional distractors are introduced. The augmentations during training increased partial success rates to \SI{50}{\percent}, but fully complete insertions remain few (\SI{5}{\percent}). Substantially more diverse data is required to achieve robustness to various object placements.

\input{table_generalisation}

\vspace{1mm}
\subsection{What Matters During Data Collection?}

We analyze the key parameters of our autonomous data-acquisition scheme. We use a MuJoCo~\citep{todorov2012mujoco} simulated setup of the LEGO tasks, train policies on three different seeds, and report the results on 1000 policy rollouts.

\textit{Successful policies require dense data coverage at the contact-rich segment (\autoref{fig:results_datacollection} left+middle).} Policies trained on data with little exploration ($p \leq 0.3$) achieve low success rates. Little exploration results in short trajectories that yield little training data (left). Note that $p=0$ is close to imitation learning data. On the other hand, pure exploration ($p=1$) leads to a low data-collection SR and thus to unsuccessful policies. In practice, rollouts must be sufficiently long while maintaining a high success rate, which we observe for $0.6 \leq p \leq 0.85$. The time spent in the critical segment per rollout is decisive, which requires DRL to learn from suboptimal trajectories.

\textit{Scaling of success rate and dataset size with pose estimation uncertainty (\autoref{fig:results_datacollection} right).} Increasingly uncertain pose estimation expands the critical segment of the task. In simulation, we require 400 rollouts for \SI{6}{\milli\metre} and 700 rollouts for \SI{12}{\milli\metre} uncertainty. This corresponds to \SIrange{2}{3.5}{\hour} of data collection in the real world. We tend to require more data in simulation than on hardware for the same success rates, possibly because of inconsistent simulation physics during contact.

\section{Conclusion}
\label{sec:conclusion}

Robot data is expensive, and we should consider how to collect and use it efficiently.  While clearly more robot data is needed overall, we argue for focusing collection on tasks and task segments that are otherwise difficult to solve. By adopting a modular approach with targeted data collection, we maintain high success rates for OOD scenarios, demonstrating that focused use of robot data and structured task decomposition are key enablers of robust, generalizable manipulation while keeping data-collection effort low.

\begin{figure}[!t]
    \centering
    \includegraphics[width=\columnwidth]{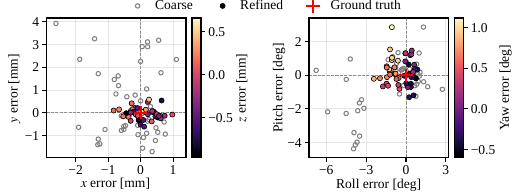}\par
    \vspace{2pt}
    \includegraphics[width=0.49\columnwidth]{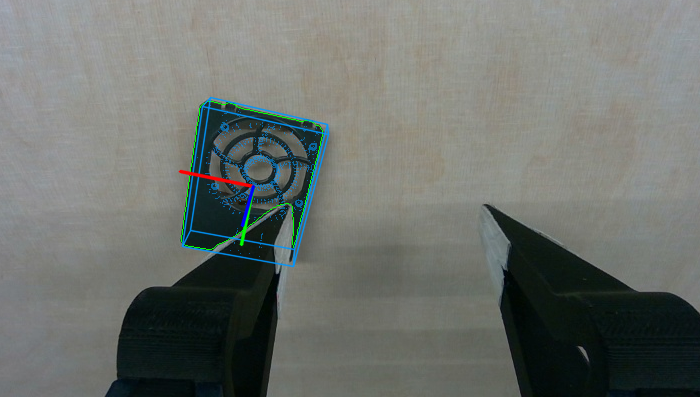}%
    \hfill
    \includegraphics[width=0.49\columnwidth]{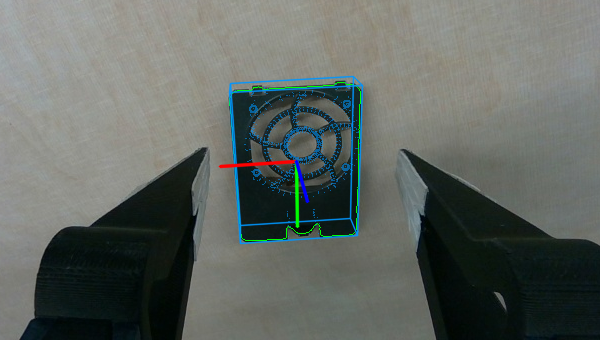}
    \caption{\textbf{Pose estimation.} \textbf{Top}: Translation (left) and rotation (right) errors of coarse and fine estimates. \textbf{Bottom}: Qualitative coarse (left) and fine (right) estimate. For the fine estimates, all estimates fall within a range of $\pm$\SI{1}{\milli\metre} and $\pm$\SI{3.3}{\degree}.}
    \label{fig:app_pe_summary}
\end{figure}

\appendix[Evaluation of Pose Estimation]
\label{sec:app_pe}
We evaluate the pose estimation (FoundationPose~\citep{wen2024foundationposeunified6dpose}) quantitatively on the fan cover task. For the evaluation protocol, we placed the fan cover at a fixed position in the scene. We then run pose estimation from a distance of $\approx \SI{30}{\centi\meter}$ (coarse) and $\approx \SI{10}{\centi\meter}$ (fine) each 50 times from different end-effector positions. Quantitative and qualitative results are reported in~\autoref{fig:app_pe_summary}. For the fine estimates, all estimates fall within a range of $\pm$\SI{1}{\milli\metre} and $\pm$\SI{3.3}{\degree}.

The critical segment is defined by the uncertainty in the pose estimation---effectively, we correct for this uncertainty using a local policy trained on robot data. In practice, we choose the critical segment to be larger than the expected uncertainty.

\renewcommand*{\bibfont}{\footnotesize}
\printbibliography

\input{appendix}

\end{document}

%% file: task_images.tex
\newlength{\taskimgw}\setlength{\taskimgw}{0.235\textwidth}
\newlength{\taskimgh}\setlength{\taskimgh}{\dimexpr\taskimgw*2/3\relax} %
\newlength{\taskimggap}\setlength{\taskimggap}{0.01\textwidth}
\newcommand{\taskimg}[5][-1]{%
    \begin{tikzpicture}
      \path[use as bounding box] (0,0) rectangle (\taskimgw, \taskimgh);
      \clip (0,0) rectangle (\taskimgw, \taskimgh);
      \node[anchor=center, inner sep=0, rotate=#1]
        at ({(0.5+#4)*\taskimgw}, {0.5*\taskimgh+#5*\taskimgw}) {%
        \includegraphics[width=#3\taskimgw]{#2}};
    \end{tikzpicture}%
}

\begin{figure*}[!t]
    \centering
    \noindent
    \begin{minipage}[c]{\dimexpr2\taskimgw+\taskimggap\relax}
    \taskimg{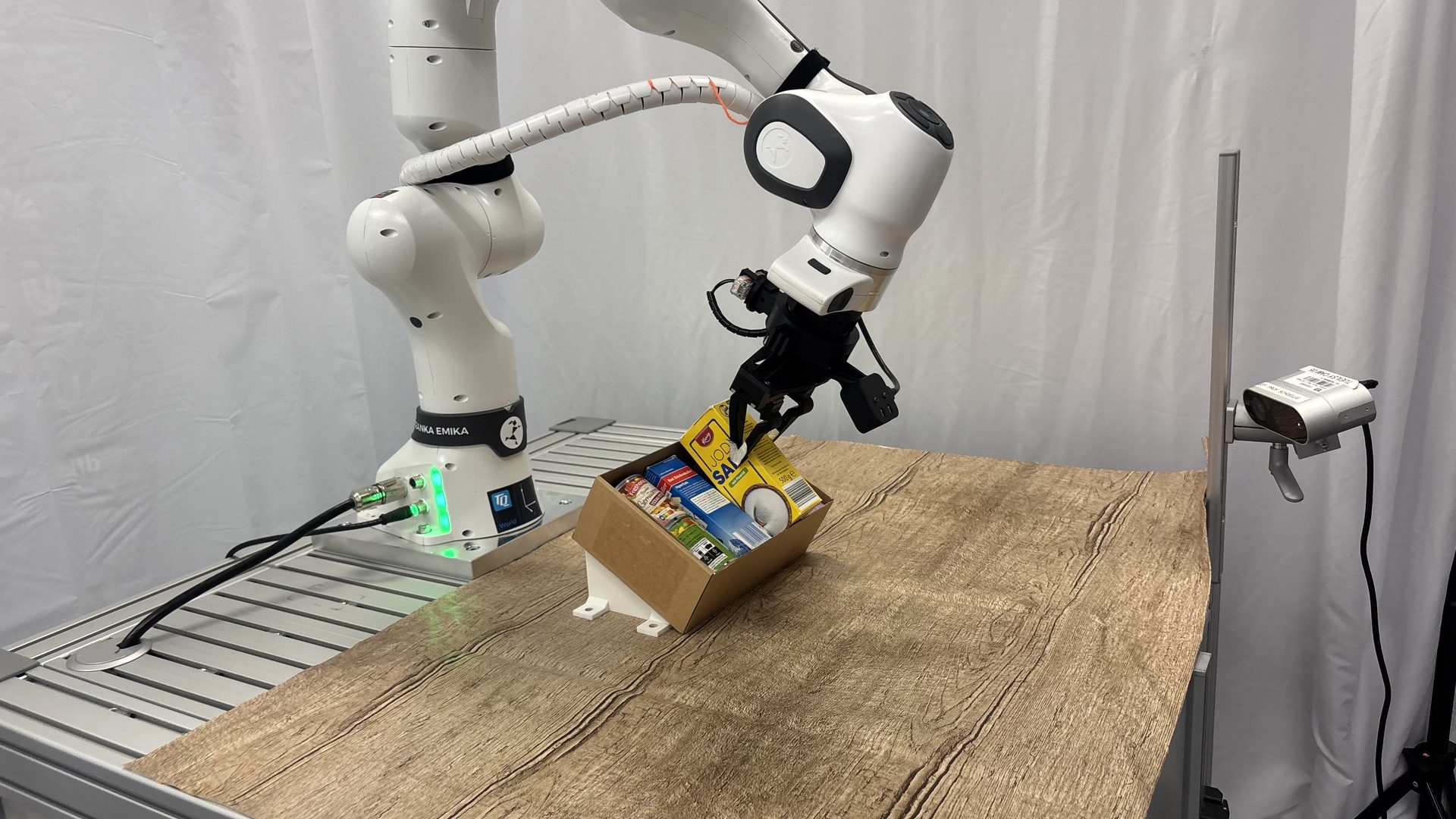}{2.3415}{0.0122}{0.0976}%
    \hfill
    \taskimg{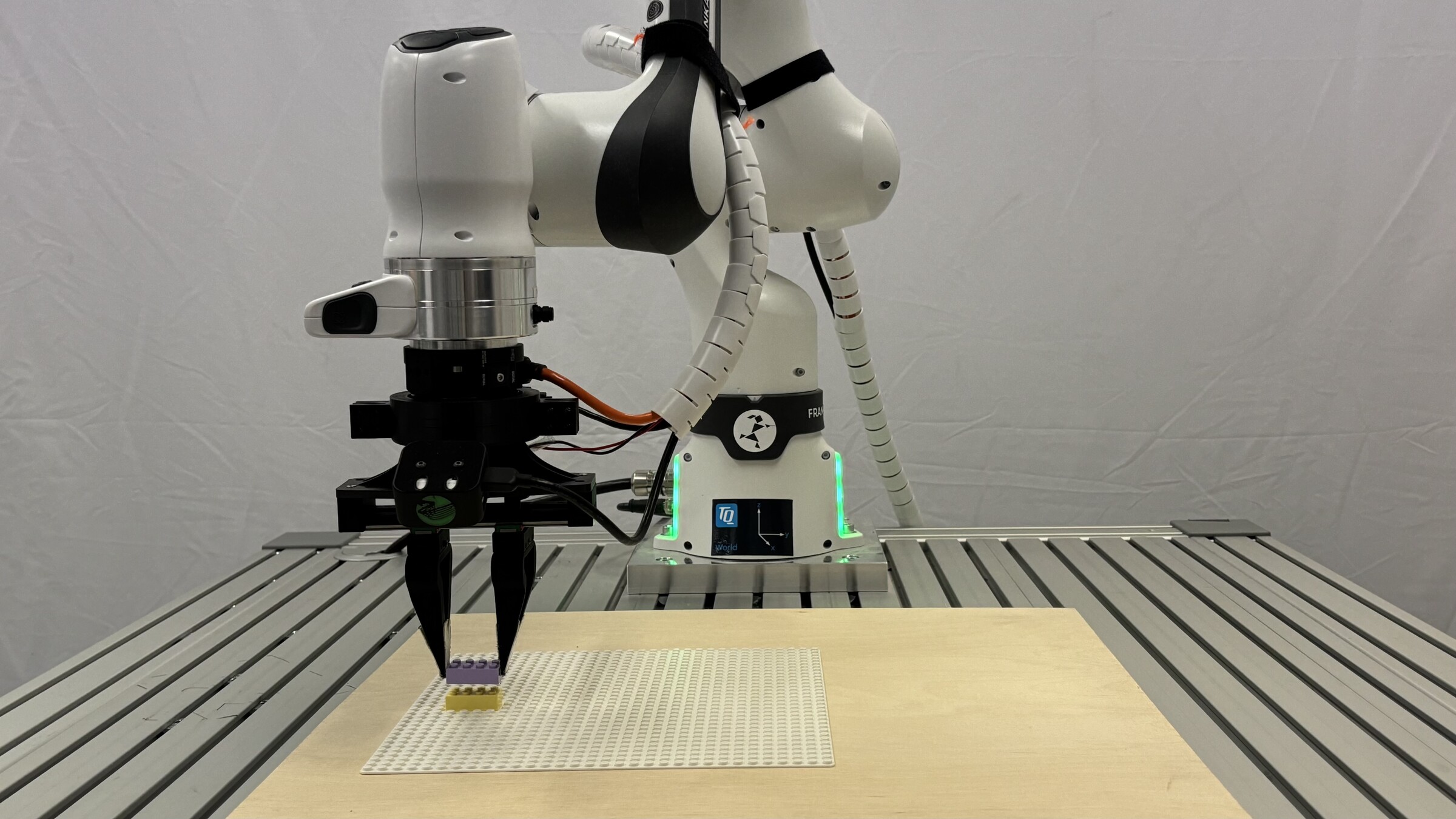}{2.7907}{0.4070}{0.3314}%

    \vspace{\taskimggap}

    \taskimg{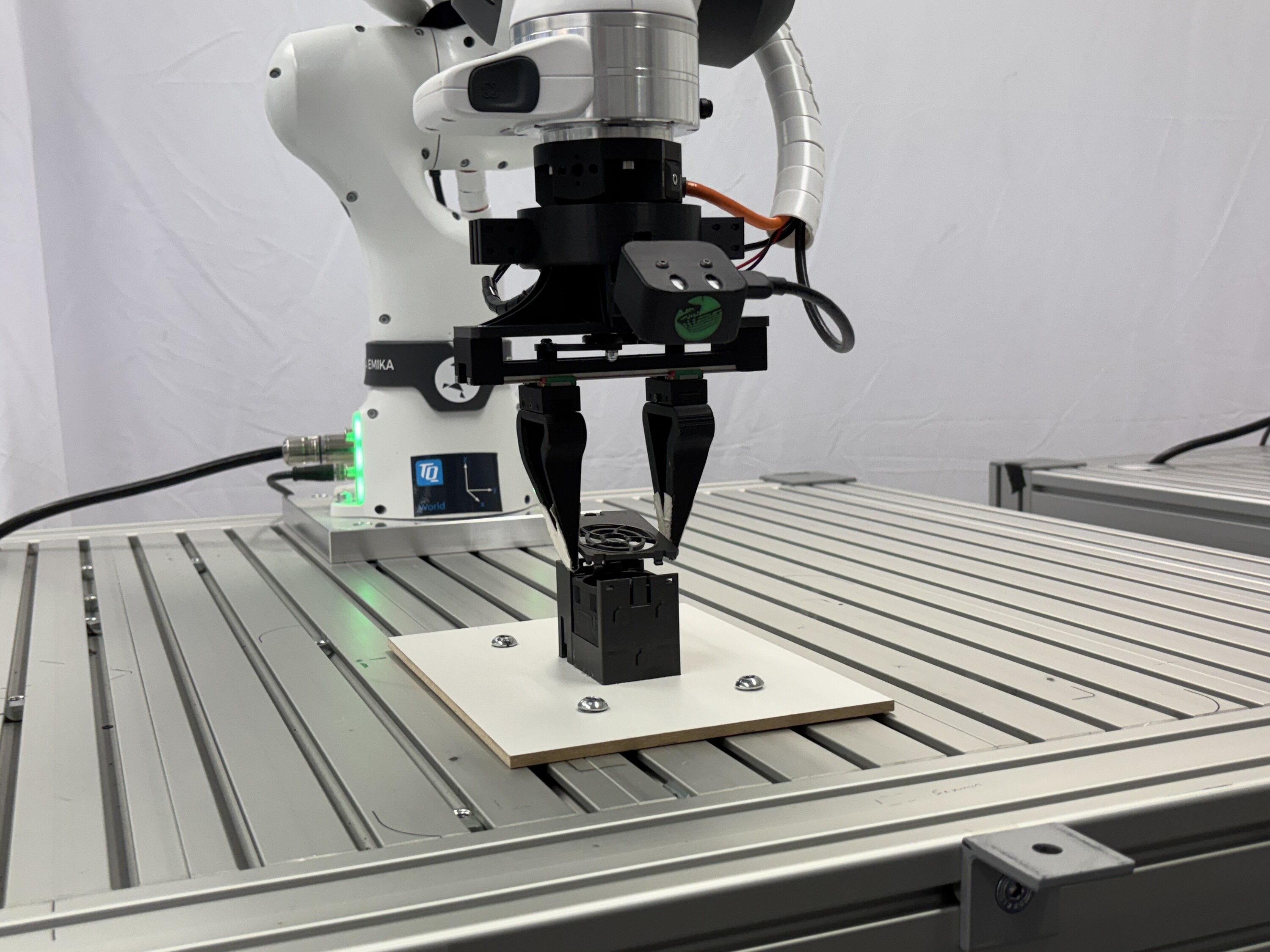}{3.0000}{0.0500}{0.1950}%
    \hfill
    \taskimg{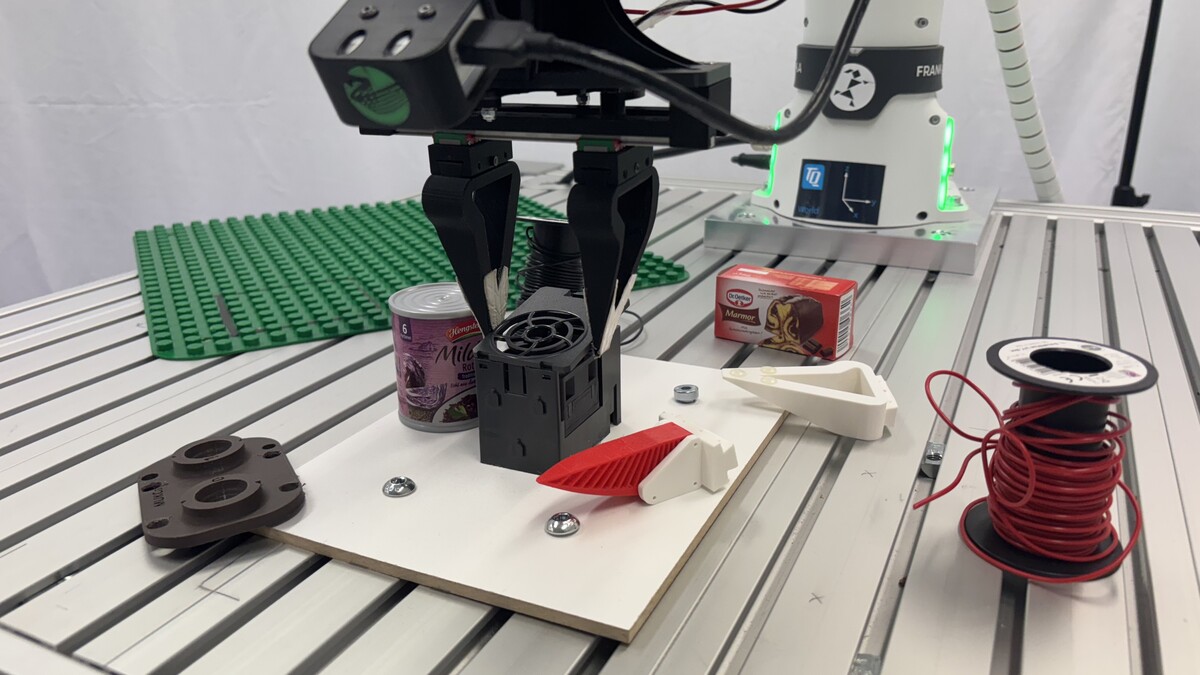}{1.5385}{-0.0128}{0.0288}%
    \end{minipage}%
    \hfill
    $\vcenter{\hbox{\rule{0.4pt}{\dimexpr2\taskimgh+\taskimggap\relax}}}$%
    \hfill
    \begin{minipage}[c]{\taskimgw}
    \taskimg[0]{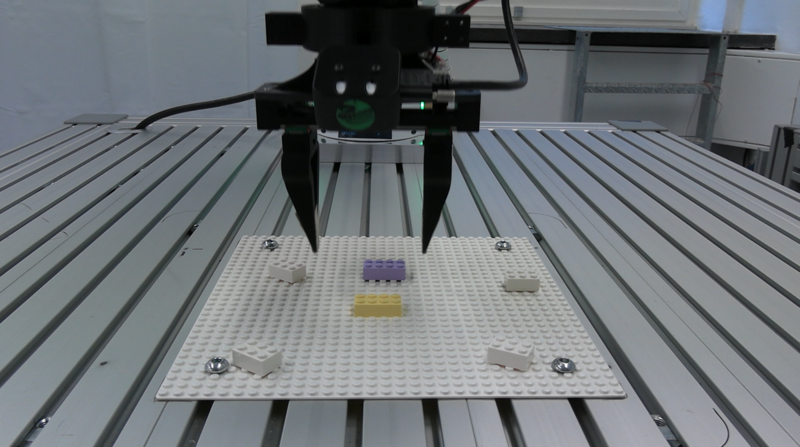}{1.3}{0}{0}%

    \vspace{\taskimggap}

    \taskimg[0]{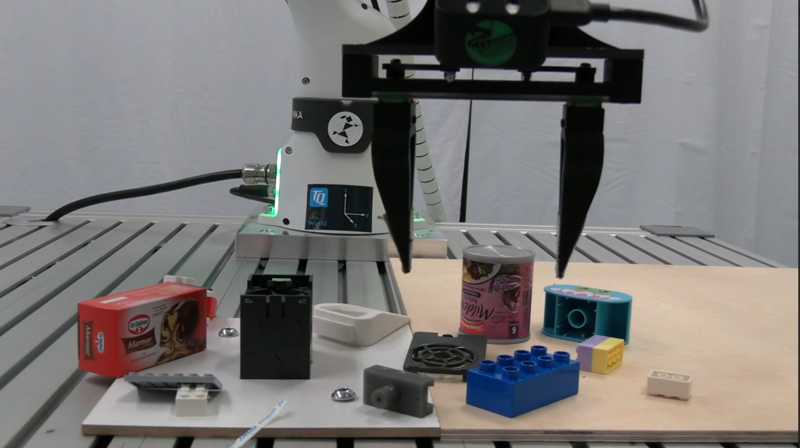}{1.3}{0}{0}%
    \end{minipage}%
    \hfill
    $\vcenter{\hbox{\rule{0.4pt}{\dimexpr2\taskimgh+\taskimggap\relax}}}$%
    \hfill
    \begin{minipage}[c]{\taskimgw}
    \taskimg[0]{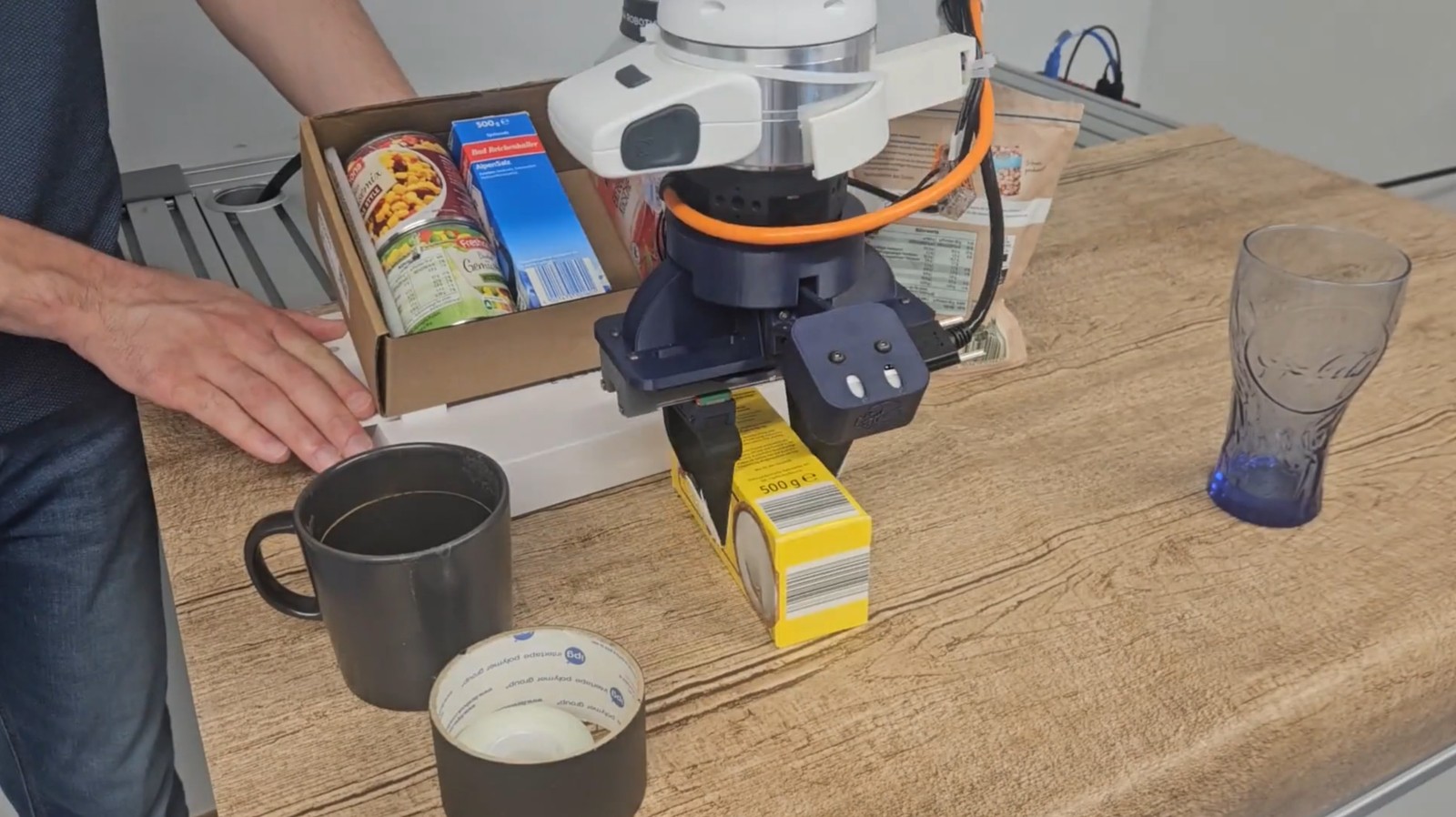}{1.3}{0}{0}%

    \vspace{\taskimggap}

    \taskimg[0]{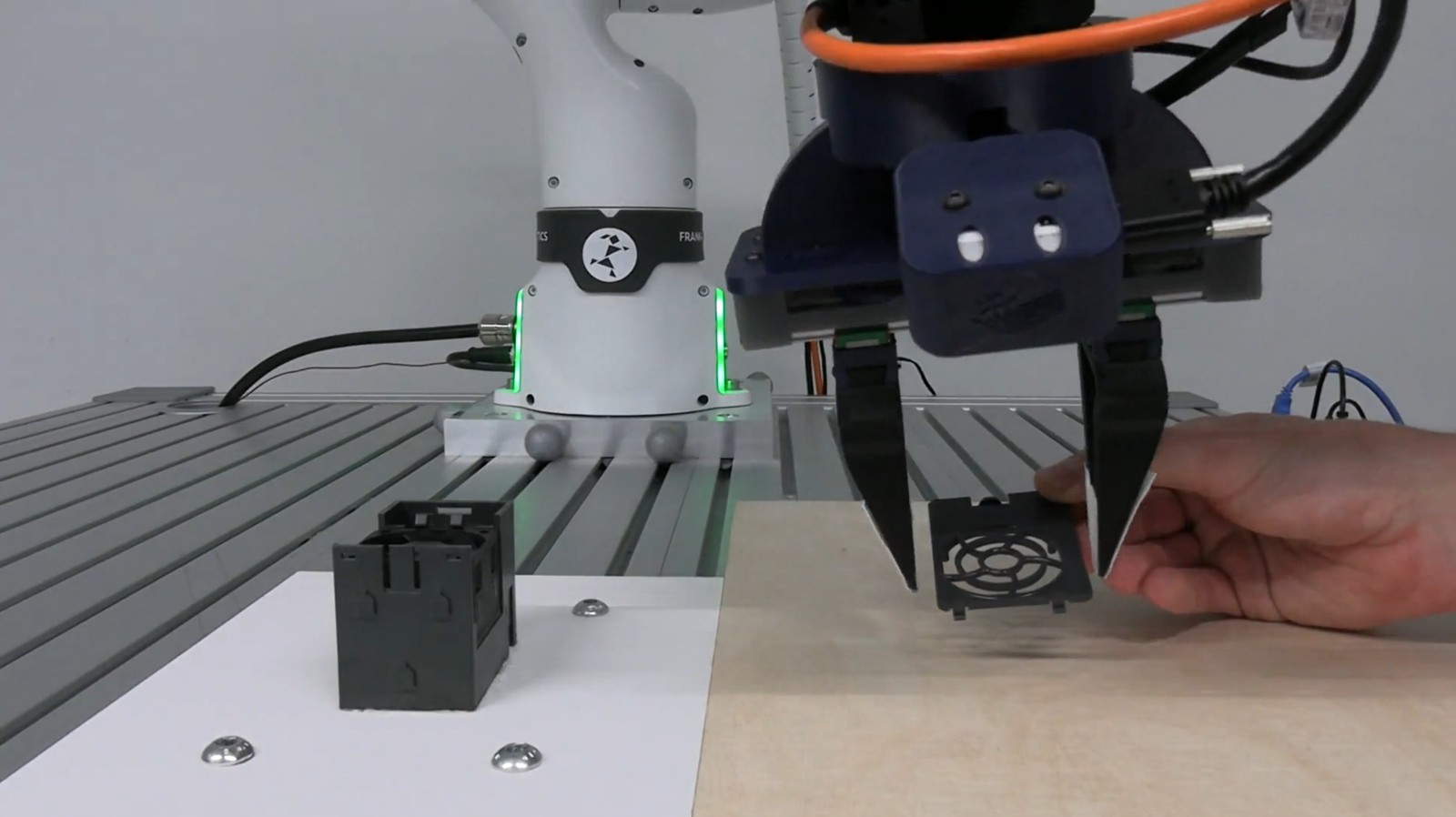}{1.3}{0}{0}%
    \end{minipage}

    \caption{\textbf{Tasks.} (\textbf{Left}) We evaluate our method on four challenging real world tasks: shelf stocking, LEGO stacking, mounting two PBT/PC-molded parts (fan cover), and a hard version of the fan cover task that we train on scene distractors, similar objects, and changing ground surfaces. (\textbf{Middle}) Example settings for out-of-distribution scenarios: for each task, we evaluate under slight distractors (top), severe distractors (bottom), and background changes (not shown here). (\textbf{Right}) Additional scenarios for which we test our method are in the supplementary video, such as picking from a human hand.}
    \label{fig:tasks_images}
\end{figure*}

%% file: table_time_vertical.tex
\begin{table}[!t]
\centering
\begin{threeparttable}
\caption{Data collection}
\label{tab:data_efficiency}
\footnotesize
\setlength{\tabcolsep}{4pt}
\begin{tabular*}{\columnwidth}{@{\extracolsep{\fill}} l l c c c @{}}
\toprule
\textbf{Task} & \textbf{Method} & \textbf{\makecell{\# Episodes\\(Failures\tnote{1})}} & \textbf{\makecell{\# Data-\\points}} & \textbf{\makecell{Wall-clock\\time}} \\
\midrule
\multirow{4}{*}{\makecell[l]{Shelf\\stocking}}
 & Teleop                     & 100 (22)  & 24k & \SI{2.3}{\hour} \\
 & Teleop (Novice)\tnote{2}   & 100 (18)  & 34k & \SI{2.3}{\hour} \\
 & HIL-SERL                   & 346 (133) & 27k & \SI{3.2}{\hour} \\
 & \textbf{Ours}              & \textbf{300} & \textbf{27k} & \textbf{\SI{2}{\hour}} \\
\cmidrule{1-5}
\multirow{4}{*}{\makecell[l]{LEGO\\stacking}}
 & Teleop                     & 100 (30)  & 33k & \SI{2.5}{\hour} \\
 & Teleop (Novice)\tnote{2}   & 100 (27)  & 42k & \SI{2.6}{\hour} \\
 & HIL-SERL                   & 495 (221) & 32k & \SI{3.2}{\hour} \\
 & \textbf{Ours}              & \textbf{500} & \textbf{30k} & \textbf{\SI{2.5}{\hour}} \\
\cmidrule{1-5}
\multirow{2}{*}{Fan Cover}
 & Teleop                     & 100 (50)  & 39k & \SI{2.75}{\hour} \\
 & \textbf{Ours}              & \textbf{300} & \textbf{33k} & \textbf{\SI{2}{\hour}} \\
\cmidrule{1-5}
\multirow{2}{*}{\makecell[l]{Fan Cover\\(hard)}}
 & Teleop                     & 100 (55)  & 40k & \SI{2.5}{\hour} \\
 & \textbf{Ours}              & ---\rlap{\tnote{3}} & --- & --- \\
\bottomrule
\end{tabular*}
\begin{tablenotes}[para]
    \scriptsize
    \textsuperscript{1}Episodes that failed during teleoperation---i.e., where the teleoperator did not successfully complete the task---are indicated in brackets and were excluded from policy training. \textsuperscript{2}``Novice'' refers to an inexperienced teleoperator who has been given a couple of trials before starting data collection. All other data have been collected by an experienced operator. \textsuperscript{3}No additional data needed.
\end{tablenotes}
\end{threeparttable}
\end{table}

%% file: table_main_results.tex
\begin{table*}[!t]
    \centering
    \begin{threeparttable}
    \caption{Success Rates (Partial Success Rates\tnote{1}) [\%]}
    \label{tab:sr_results} \small
    \setlength{\tabcolsep}{0pt}
    \begin{tabular*}{\textwidth}{@{\extracolsep{\fill}} l cccccc c@{}}
        \toprule
        \textbf{Task} & \textbf{\shortstack{DiTFlow}} & \textbf{\shortstack{DiTFlow\\(Novice)\tnote{2}}} & \textbf{DP} & \textbf{\shortstack{Fine-tuned\\$\pi_{0.5}$}} & \textbf{PE \& MP\tnote{3}} & \textbf{\shortstack{HIL-\\SERL}} & \textbf{Ours} \\
        \midrule
        Shelf stocking       & 38 (96)           & 34 (80)        & 44 (100)           & 42 (92)           &    98 (98)      & 10 (24)   & \textbf{98 (100)} \\
        LEGO stacking       & 60 (88) & 22 (48)      & 44 (92) & 6 (80) & 78 (100) & 6 (22)   & \textbf{94 (100)} \\
        Fan Cover           & 30 (88) & ---    & 30 (96) & 20 (96)& 22 (94)  & --- & \textbf{96 (98)} \\
        Fan Cover (hard)    & 18 (82)           & ---      & 26 (92)           & 12 (76)          &  22 (94)            & ---   & \textbf{94 (96)}\tnote{4} \\
        \midrule
        Average  & 37 (89) & --- & 36 (95) & 20 (86) & 55 (97) & --- & \textbf{96 (99)} \\
        \bottomrule
    \end{tabular*}
    \begin{tablenotes}[para]
        \scriptsize
        We roll out each policy for 50 trials. \textsuperscript{1}Partial success is when the objects are placed correctly but not fully inserted or assembled. \textsuperscript{2}``Novice'' refers to a policy trained on data collected by a novice teleoperator. All other data have been collected by an experienced operator. \textsuperscript{3}Pose Estimation \& Motion Planning. \textsuperscript{4}We deployed the same policy as for the simple version of the task without additional data collection.
    \end{tablenotes}
    \end{threeparttable}
\end{table*}

%% file: table_generalisation.tex
\begin{table}[!t]
    \centering
    \begin{threeparttable}
    \caption{Success Rates (Partial Success Rates\textsuperscript{1}) [\%] for out-of-distribution scenarios}
    \label{tab:sr_results_generalization}
    \small
    \setlength{\tabcolsep}{0pt}
    \begin{tabular*}{\columnwidth}{@{\extracolsep{\fill}} l cccc c@{}}
        \toprule
        \textbf{Task} & \textbf{\shortstack{DiTFlow}} & \textbf{DP} & \textbf{\shortstack{Fine-tuned\\$\pi_{0.5}$}} & \textbf{\shortstack{HIL-\\SERL}} & \textbf{Ours} \\
        \midrule
        Shelf stocking       & 20 (40) & 25 (45) & 10 (65) & 0 (5) & \textbf{90 (100)} \\
        LEGO stacking        & 10 (10)   & 10 (25)  & 0 (0)   & 0 (0) & \textbf{85 (95)} \\
        Fan Cover            & 0 (30)  & 0 (20)  & 0 (50)  & --- & \textbf{90 (95)} \\
        Fan Cover (hard)     & 0 (50)  & 5 (50)  & 5 (50)  & --- & \textbf{95 (95)}\tnote{2} \\
        \bottomrule
    \end{tabular*}
    \begin{tablenotes}[para]
        \scriptsize
        We roll out each policy for 20 trials. \textsuperscript{1}Partial success is when the objects are placed correctly but not fully inserted or assembled. \textsuperscript{2}We deployed the same policy as for the simple version of the task---without additional data collection for this task.
    \end{tablenotes}
    \end{threeparttable}
\end{table}

%% file: appendix.tex
\FloatBarrier

\FloatBarrier